\documentclass[10pt,twocolumn,letterpaper]{article}

\usepackage{3dv}
\usepackage{times}
\usepackage{epsfig}
\usepackage{graphicx}
\usepackage{subcaption}
\usepackage{amsmath}
\usepackage{amssymb}
\usepackage{multirow}
\usepackage{color, colortbl}
\usepackage{bm}

\definecolor{Gray}{gray}{0.9}
\newcolumntype{g}{>{\columncolor{Gray}}c}

\makeatletter
\def\thickhline{%
  \noalign{\ifnum0=`}\fi\hrule \@height \thickarrayrulewidth \futurelet
   \reserved@a\@xthickhline}
\def\@xthickhline{\ifx\reserved@a\thickhline
               \vskip\doublerulesep
               \vskip-\thickarrayrulewidth
             \fi
      \ifnum0=`{\fi}}
\makeatother

\newlength{\thickarrayrulewidth}
\newcommand{\vect}[1]{\mathbf{#1}}

\newcommand\blfootnote[1]{%
  \begingroup
  \renewcommand\thefootnote{}\footnote{#1}%
  \addtocounter{footnote}{-1}%
  \endgroup
}

\usepackage[pagebackref=true,breaklinks=true,colorlinks,bookmarks=false]{hyperref}

\threedvfinalcopy 

\def\threedvPaperID{233} 

\ifthreedvfinal\fi

\begin{document}

\title{3DVNet: Multi-View Depth Prediction and Volumetric Refinement}

\author{
Alexander Rich
\and
Noah Stier
\and
Pradeep Sen
\and
Tobias Höllerer
}
\date{{\tt\small\{anrich, noahstier, psen, thollerer\}@ucsb.edu}}
\affiliation{University of California, Santa Barbara}  

\maketitle
\thispagestyle{empty}
\global\csname @topnum\endcsname 0
\global\csname @botnum\endcsname 0

\begin{abstract}
We present 3DVNet, a novel multi-view stereo (MVS) depth-prediction method that combines the advantages of previous depth-based and volumetric MVS approaches.
Our key idea is the use of a 3D scene-modeling network that iteratively updates a set of coarse depth predictions, resulting in highly accurate predictions which agree on the underlying scene geometry.
Unlike existing depth-prediction techniques, our method uses a volumetric 3D convolutional neural network (CNN) that operates in world space on all depth maps jointly.
The network can therefore learn meaningful scene-level priors.
Furthermore, unlike existing volumetric MVS techniques, our 3D CNN operates on a feature-augmented point cloud, allowing for effective aggregation of multi-view information and flexible iterative refinement of depth maps.
Experimental results show our method exceeds state-of-the-art accuracy in both depth prediction and 3D reconstruction metrics on the ScanNet dataset, as well as a selection of scenes from the TUM-RGBD and ICL-NUIM datasets.
This shows that our method is both effective and generalizes to new settings.
\end{abstract}

\section{Introduction}

\begin{figure}
\begin{center}
   \includegraphics[width=\columnwidth]{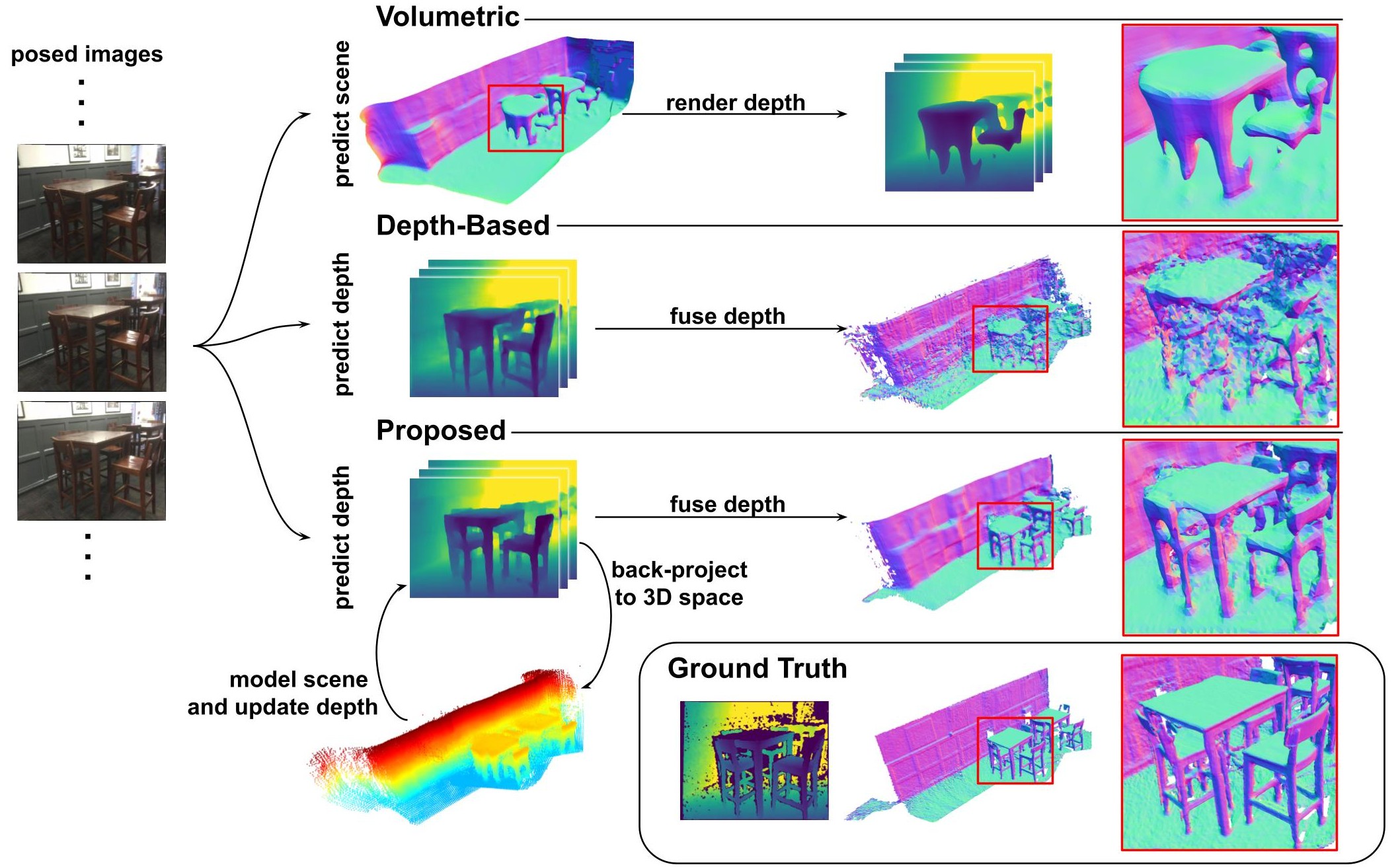}
\end{center}
\vspace{-0.6cm}
\begin{small}
\caption{Volumetric methods lack local detail while depth-based methods lack global coherence. Our method cyclically predicts depth, back-projects into 3D space, volumetrically models geometry, and updates all depth predictions to match, resulting in local detail \textit{and} global coherence.}
\label{fig:diffs}
\end{small}
\end{figure}

\blfootnote{\url{https://github.com/alexrich021/3dvnet}}Multi-view stereo (MVS) is a central problem in computer vision with applications from augmented reality to autonomous navigation.
In MVS, the goal is to reconstruct a scene using only posed RGB images as input.
This reconstruction can take many forms, from voxelized occupancy or truncated signed distance fields (TSDFs), to per-frame depth prediction, the focus of this paper.
In recent years, MVS methods based on deep learning \cite{chen2019ptmvs, duzceker2020dvmvs, hou2019gpmvs, im2019dps, luo2019pmvs, murez2020atlas, sinha2020deltas, sun2021neucon, mvdepthnet, yao2018mvs, yao2019rmvs, yi2020pyramid, yu2020fmvs} have surpassed traditional MVS methods \cite{galliani2015gaupuma, schoenberger2016mvs} on numerous benchmark datasets \cite{dai2017scannet, jensen2014large, knapitsch2017tanks}.
In this work, we consider these methods as falling into two categories, depth estimation and volumetric reconstruction, each with advantages and disadvantages.

The most recent learning methods in depth estimation use deep features to perform dense multi-view matching robust to large environmental lighting changes and textureless or specular surfaces, among other things.
These methods take advantage of well researched multi-view aggregation techniques and the flexibility of depth as an output modality.
They formulate explicit multi-view matching costs and include iterative refinement layers in which a network predicts a small depth offset between an initial prediction and the ground truth depth map \cite{chen2019ptmvs, yu2020fmvs}.
While these techniques have been successful for depth prediction, most are constrained to making independent, per-frame predictions.
This results in predictions that do not agree on the underlying 3D geometry of the scene.
Those that do make joint predictions across multiple frames use either regularization constraints \cite{hou2019gpmvs} or recurrent neural networks (RNNs) \cite{duzceker2020dvmvs} to encourage frames close in pose space to make similar predictions.
However, these methods do not directly operate on a unified 3D scene representation, and their resulting reconstructions lack global coherence (see Fig.~\ref{fig:diffs}).

Meanwhile, volumetric techniques operate directly on a unified 3D scene representation by back-projecting and aggregating 2D features into a 3D voxel grid and using a 3D convolutional neural network (CNN) to regress a voxelized parameter, often a TSDF.
These methods benefit from the use of 3D CNNs and naturally produce highly coherent 3D reconstructions and accurate depth predictions.
However, they do not explicitly formulate a multi-view matching cost like depth-based methods, generally averaging deep features from different views to populate the 3D voxel grid.
This results in overly-smooth output meshes (see Fig.~\ref{fig:diffs}).

In this paper, we propose \textit{3DVNet}, an end-to-end differentiable method for learned multi-view depth prediction that leverages the advantages of both volumetric scene modeling and depth-based multi-view matching and refinement.
The key idea behind our method is the use of a 3D scene-modeling network which outputs a multi-scale volumetric encoding of the scene.
This encoding is used with a modified PointFlow algorithm \cite{chen2019ptmvs} to iteratively update a set of initial coarse depth predictions, resulting in predictions that agree on the underlying scene geometry.

Our 3D network operates on all depth predictions at once, and extracts meaningful, scene-level priors similar to volumetric MVS methods.
However, the 3D network operates on features aggregated using depth-based multi-view matching and can be used iteratively to update depth maps.
In this way, we combine the advantages of the two separate classes of techniques.
Because of this, 3DVNet exceeds state-of-the-art results on ScanNet \cite{dai2017scannet} in nearly all depth map prediction \textit{and} 3D reconstruction metrics when compared with the current best depth and volumetric baselines.
Furthermore, we show our method generalizes to other real and synthetic datasets \cite{handa2014icl-nuim, sturm2012tum-rgbd}, again exceeding the best results on nearly all metrics.
Our contributions are as follows:
\begin{enumerate}
    \item We present a 3D scene-modeling network which outputs a volumetric scene encoding, and show its effectiveness for iterative depth residual prediction.
    \item We modify PointFlow~\cite{chen2019ptmvs}, an existing method for depth map residual predictions, to use our volumetric scene encoding.
    \item We design 3DVNet, a full MVS pipeline, using our 3D scene-modeling network and PointFlow refinement.
\end{enumerate}

\section{Related Works}
We cover MVS methods using deep learning, categorizing them as either depth-prediction methods or volumetric methods.
Our method falls into the first category, but is very much inspired by volumetric techniques.

\textbf{Depth-Prediction MVS Methods:} With some notable exceptions \cite{sinha2020deltas, yang2021mvs2d}, nearly all depth-prediction methods follow a similar paradigm: (1) they construct a plane sweep cost volume on a reference image's camera frustum, (2) they fill the volume with deep features using a cost function that operates on source and reference image features, (3)~they use a network to predict depth from this cost volume.
Most methods differ in their cost metric used to construct the volume.
Many cost metrics exist, including per-channel variance of deep features \cite{yao2018mvs, yao2019rmvs}, learned aggregation using a network \cite{luo2019pmvs, yi2020pyramid}, concatenation of deep features \cite{im2019dps}, the dot product of deep features \cite{duzceker2020dvmvs}, and absolute intensity difference of raw image RGB values \cite{hou2019gpmvs, mvdepthnet}.
We find per-channel variance \cite{yao2018mvs} to be the most commonly used cost metric, and adopt it in our system.

The choice of cost aggregation method results in either a vectorized matching cost and thus a 4D cost volume \cite{chen2019ptmvs, im2019dps, luo2019pmvs, yao2018mvs, yao2019rmvs, yi2020pyramid, yu2020fmvs} or a scalar matching cost and thus a 3D cost volume \cite{duzceker2020dvmvs, hou2019gpmvs, mvdepthnet}.
Methods with 4D cost volumes generally require 3D networks for processing, while 3D cost volumes can be processed with a 2D U-Net-style \cite{ronneberger2015unet} encoder-decoder architecture.
Some methods operate on the deep features at the bottleneck of this U-Net to make joint depth predictions for all $N$ frames or a subset of frames in a given scene.
This is similar to our proposed method, and we highlight the differences.

GPMVS \cite{hou2019gpmvs} uses a Gaussian Process (GP) constraint conditioned on pose distance to regularize these deep features.
This GP constraint only operates on deep features and assumes Gaussian priors.
In contrast, we directly \textit{learn} priors from predicted depth maps and explicitly predict depth residuals to modify depth maps to match.
DV-MVS \cite{duzceker2020dvmvs} introduces an RNN to propagate information from the deep features in frame $t - 1$ to frame $t$ given an ordered sequence of frames.
While they do propagate this information in a geometrically plausible way, the RNN operates only on deep features similar to GPMVS.
Furthermore, the RNN never considers all frames jointly like our method.

Similar to our method, some networks iteratively predict a residual to refine an initial depth prediction \cite{chen2019ptmvs, yu2020fmvs}.
We specifically highlight Point-MVSNet \cite{chen2019ptmvs}, which introduces PointFlow, a point cloud learning method for residual prediction. 
Our method is very much inspired by this work. We briefly describe the differences.

In their work, they operate on a point cloud back-projected from a \textit{single} depth map and augmented with additional points.
Features are extracted from this point cloud using point cloud learning techniques and used in their PointFlow module for residual prediction.
Crucially, these features do not come from a unified 3D representation of the scene.
Thus the residual prediction is only conditioned on information local to the individual depth prediction and not global scene information.
In contrast, our variation of PointFlow uses our volumetric scene model to condition residual prediction on information from \textit{all} depth maps.
For an in depth discussion of differences, see Sec.~\ref{sec:refinement}.

\textbf{Volumetric MVS Methods:} In volumetric MVS, the goal is to directly regress a global volumetric representation of the scene, generally a TSDF volume.
We highlight two methods which inspired our work.
Atlas \cite{murez2020atlas} back-projects rays of deep features extracted from images into a global voxel grid, pools features from multiple views using a running average, then directly regress a TSDF in a coarse-to-fine fashion using a 3D U-Net.
NeuralRecon~\cite{sun2021neucon} improves on the memory consumption and run-time of Atlas by reconstructing local fragments using the most recent keyframes, then fusing the local fragments to a global volume using an RNN.
The reconstructions these methods produce are pleasing.
However, both construct feature volumes using averaging in a single forward pass, which we believe is non-optimal.
In contrast, our depth-based method allows us to construct a feature volume using multi-view matching features and perform iterative refinement.

\begin{figure*}[h]
\begin{center}
   \includegraphics[width=0.9\linewidth]{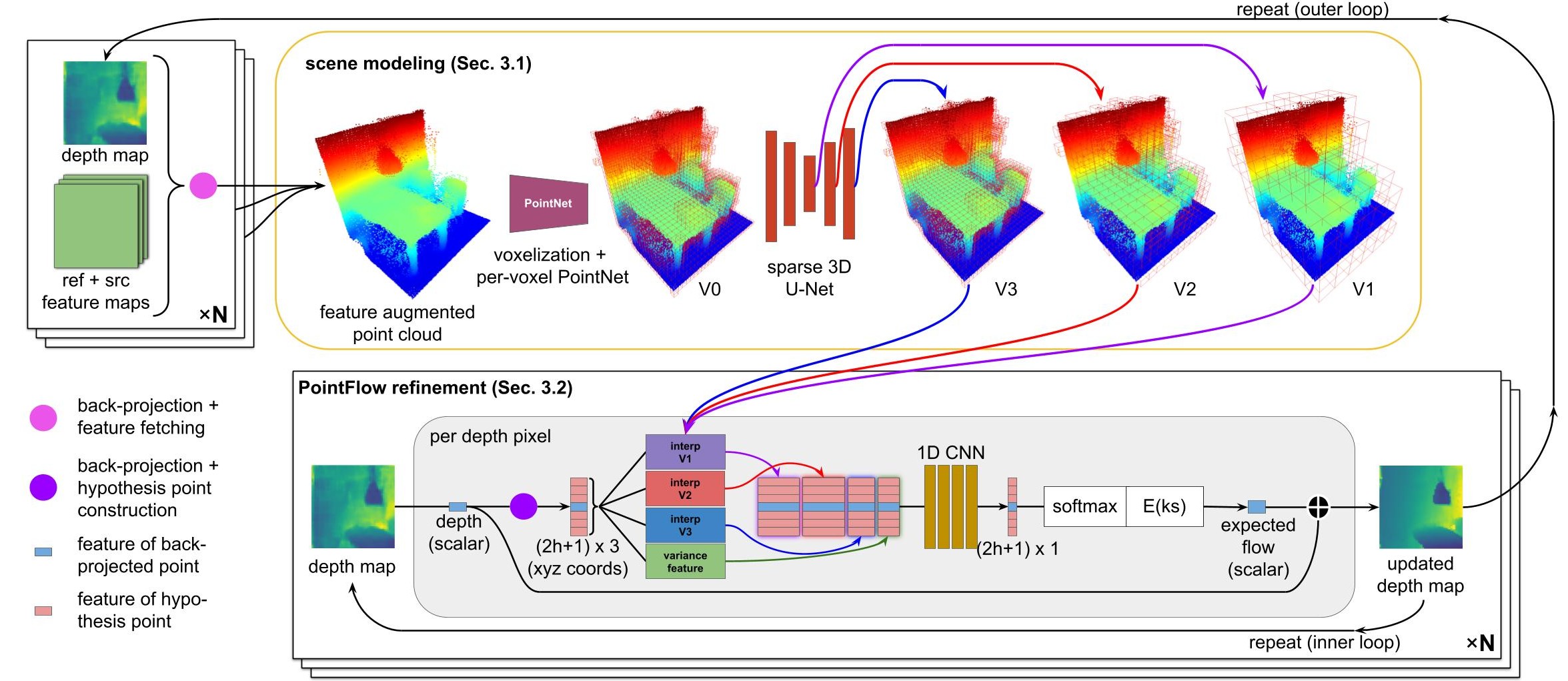}
\end{center}
\vspace{-0.6cm}
\caption{Our novel 3D scene modeling and refinement method first constructs a multi-scale volumetric scene encoding from a set of $N$ input depth maps with corresponding feature maps. It then uses that encoding in a variation of the PointFlow algorithm~\cite{chen2019ptmvs} to predict a residual for each of the $N$ depth maps. The full method can be run in a nested for-loop fashion, predicting multiple residuals per depth map in the inner loop and running scene modeling in the outer loop.}
\label{fig:main}
\end{figure*}

\section{Methods}

Our method takes as input $N$ images, denoted $\{\vect{I}_n\}$, $n=1, \dots, N$ with corresponding known extrinsic and intrinsic camera parameters.
Our goal is to predict $N$ depth maps $\{\vect{D}_n\}$ corresponding to the $N$ images.
As a pre-processing step, we define for every image $\vect{I}_n$ a set of $M$ indices $\{s_1, \dots, s_M\}$ pointing to which images to use as source images for depth prediction,
and append the reference index to form the set $\vect{S}_n = \{n, s_1, \dots, s_M\}$.

Our pipeline is as follows.
First, a small depth-prediction network is used to independently predict initial coarse depth maps $\{\vect{D}_n^0\}$ for every frame $\{\vect{I}_n\}$ using extracted image features $\{\vect{F}_n\}$ (Sec.~\ref{sec:3dvnet}).
Second, we back-project our $N$ initial depth maps to form a joint point cloud \mbox{$\vect{X} \subset \mathbb{R}^3$} (Sec~\ref{sec:3d-network}).
Because each point $\vect{p} \in \vect{X}$ is associated with one depth map $\vect{D}_n^{0}$ that has associated feature maps \mbox{$\{\vect{F}_s : s \in \vect{S}_n\}$}, we can augment it with a multi-view matching feature aggregated from those feature maps.
Third, our 3D scene-modeling network takes as input this feature-rich point cloud and outputs a multi-scale scene encoding $\vect{V}_1, \vect{V}_2, \vect{V}_3$ (Sec.~\ref{sec:3d-network}).
Fourth, we update each depth map to match this scene encoding using a modified PointFlow algorithm, resulting in highly coherent depth maps and thus highly coherent reconstructions (Sec.~\ref{sec:refinement}).
Steps 2-4 can be run in a nested for-loop, with steps 2 and 3 run in the outer loop to generate updated scene models with the current depth maps and step 4 run in the inner loop to refine depth maps with the current scene model.
We denote the updated depth map after $l_o$ outer loop iterations of scene modeling and $l_i$ inner loop iterations of updating as $\vect{D}_n^{(l_o, l_i)}$.
Finally, we upsample the resulting refined depth maps to the size of the original image in a coarse-to-fine manner, guided by deep features and the original image, to arrive at final predictions $\{\vect{D}_n\}$ for every image $\{\vect{I}_n\}$ (Sec.~\ref{sec:3dvnet}).

\subsection{3D Scene Modeling} \label{sec:3d-network}

A visualization of our 3D scene modeling method is given in the upper half of Fig.~\ref{fig:main}.
As stated previously, our 3D scene-modeling network operates on a feature rich point cloud back-projected from $\{\vect{D}_n^0\}$ or subsequent updated depth maps.
To process this point cloud, we adopt a voxelize-then-extract approach.
We first generate a sparse 3D grid of voxels, culling voxels that do not contain depth points.
To avoid losing granular information of the point cloud, we generate a deep feature for each voxel using a per-voxel PointNet \cite{qi2017pointnet}.
The PointNet inputs are the features of each depth point in the voxel as well as the 3D offset of that point to the voxel center.
Finally, we run a 3D U-Net~\cite{ronneberger2015unet} on the resulting voxelized feature volume and extract intermediate outputs at multiple resolutions.
By nature of construction, this U-Net learns meaningful, scene-level priors.
The result is a multi-scale, volumetric scene encoding.

\textbf{Point Cloud Formation:} We form our point cloud $\vect{X}~\subset~\mathbb{R}^3$ by back-projecting all depth pixels in all $N$ depth maps.
For our multi-view matching feature associated with each point $\vect{p} \in \vect{X}$, we follow existing work \cite{chen2019ptmvs, yao2018mvs} and use per-channel variance aggregation using the reference and source feature maps associated with each depth pixel.
For $\vect{p}~\in~\vect{X}$, given that $\vect{p}$ belongs to depth map $\vect{D}_n^0$, the equation for variance feature $\boldsymbol{\sigma}^2(\vect{p})$, applied per-channel, is:
\begin{equation}\label{eq:var}
    \boldsymbol{\sigma}^2(\vect{p}) = \frac{1}{|\vect{S}_n|}\sum_{s \in \vect{S}_n} \left(\vect{F}_{s}(\vect{\hat{p}}_s) - \overline{\vect{F}_{*}(\vect{\hat{p}}_*)}\right)^2
\end{equation}
where $\vect{\hat{p}}_s$ is the projection of $\vect{p}$ to feature map $\vect{F}_s$, $\vect{F}_s(\vect{\hat{p}}_s)$ is the bilinear interpolation of $\vect{F}_s$ to point $\vect{\hat{p}}_s$, and $\overline{\vect{F}_{*}(\vect{\hat{p}}_*)}$ is the average interpolated feature over all indices $s \in \vect{S}_n$.
Intuitively, if $\vect{p}$ lies on a surface it is more likely to have low variance in most feature channels in $\boldsymbol{\sigma}^2(\vect{p})$ while if it doesn't lie on a surface the variance will likely be high.

\textbf{Point Cloud Voxelization:} To form our initial feature volume, we regularly sample an initial 3D grid of points $\vect{C}$ every $r=8$ cm within the axis-aligned bounding box of point cloud $\vect{X}$ and define the voxel associated with each grid point $\vect{c} \in \vect{C}$ as the $8$ $\textrm{cm}^3$ cube with center $\vect{c}$.
We denote the set of depth points that fall within a voxel with center $\vect{c} \in \vect{C}$ as $v(\vect{c})~=~\{\vect{p} \in \vect{X}: ||\vect{c} - \vect{p}||_\infty <= \frac{r}{2} \}$.
We sparsify this grid by discarding $\vect{c} \in \vect{C}$ if no depth points lie within the associated voxel, denoting this set of grid coordinates as $\vect{\hat{C}} = \{ \vect{c} \in \vect{C} : v(\vect{c}) \neq \emptyset \}$.
For $\vect{c} \in \vect{\hat{C}}$, we produce a feature for the associated voxel using PointNet \cite{qi2017pointnet} with max pooling.
The PointNet feature for each voxel is defined as:
\begin{equation} \label{eq:pn}
    \vect{f}_v(\vect{c}) = \underset{\vect{p} \in v(\vect{c})}{\triangle} h_{\theta}\left(\textrm{concat}\left[\vect{p} - \vect{c}, \boldsymbol{\sigma}^2(\vect{p})\right]\right)
\end{equation}
where $h_{\theta}$ is a learnable multi-layer perceptron (MLP), $\textrm{concat}\left[\vect{q}, \vect{f}\right]$ indicates concatenation of the 3D coordinates $\vect{q}$ with the feature channel of $\vect{f}$ to form a feature with 3 additional channels, and $\triangle$ is the channel-wise max pooling operation.
The result of this stage is a sparse feature volume $\vect{V}_0$ with features given by Eq.~\ref{eq:pn} and coordinates $\vect{\hat{C}}$.

\textbf{Multi-Scale 3D Feature Extraction:} In this stage, we use a sparse 3D U-Net to model the underlying scene geometry.
We use a basic U-Net architecture with skip connections.
Group normalization is used throughout.
See supplementary material for a more detailed description of our architecture.
Our sparse U-Net takes as input sparse feature volume $\vect{V}_0$.
From intermediate outputs of the U-Net, we extract three scales of feature volumes $\vect{V}_1$, $\vect{V}_2$, $\vect{V}_3$ with a voxel edge length of $4r = 32$ cm, $2r = 16$ cm, and $r = 8$ cm, respectively, describing the scene.
In this way, we extract a rich, multi-scale, volumetric encoding of the scene.

\begin{figure}[t]
\begin{center}
   \includegraphics[width=\linewidth]{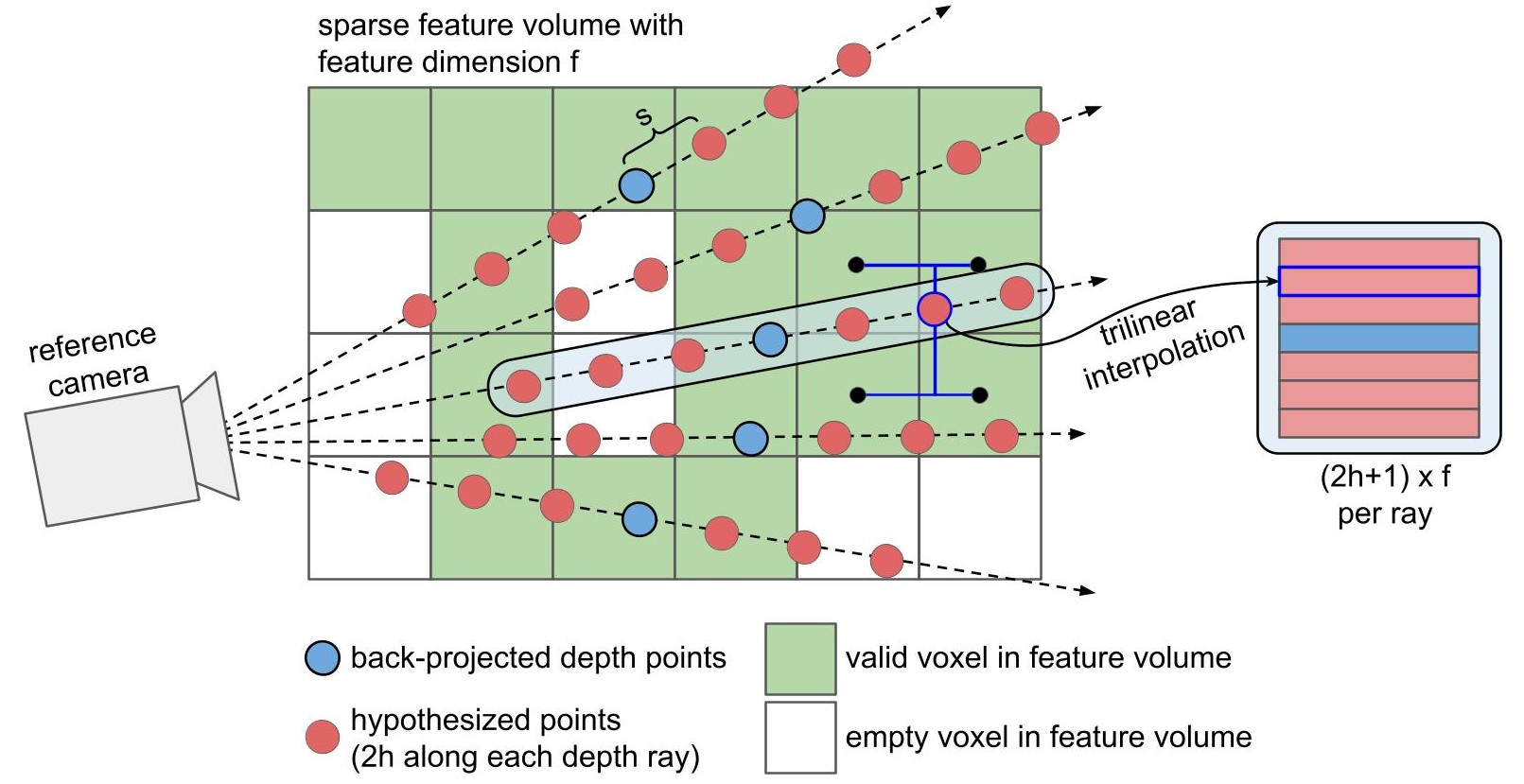}
\end{center}
\vspace{-0.6cm}
\caption{Diagram of standard PointFlow hypothesis point construction and our proposed feature generation, shown in 2D for simplicity. Feature volume in diagram corresponds to a single scale of our multi-scale scene encoding. 
Our key change from the original formulation is to generate hypothesis point features by trilinear interpolation of our volumetric scene encoding rather than edge convolutions on the point cloud from a single back-projected depth map.
}
\label{fig:interp}
\end{figure}

\subsection{PointFlow-Based Refinement} \label{sec:refinement}

\begin{figure*}[t]
\begin{center}
   \includegraphics[width=0.9\linewidth]{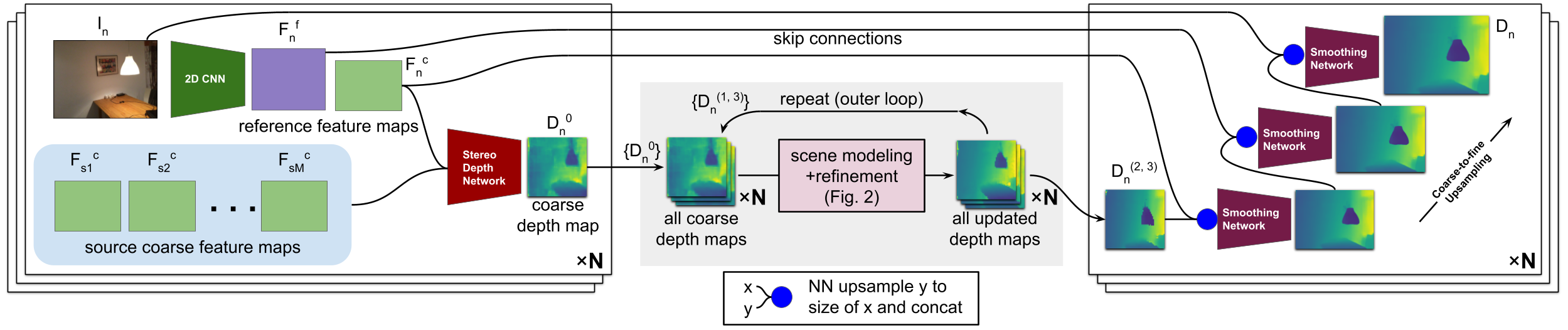}
\end{center}
\vspace{-0.6cm}
\caption{Overview of the full 3DVNet pipeline. See Secs.~\ref{sec:3d-network} and \ref{sec:refinement} for a description of our scene modeling and refinement.}
\label{fig:3dvnet}
\end{figure*}

In this stage, we use our multi-scale scene encoding $\vect{V}_1, \vect{V}_2, \vect{V}_3$ from the previous stage in a variation of the PointFlow algorithm proposed by Chen \etal~\cite{chen2019ptmvs}.
The goal is to refine our predicted depth maps to match our scene model by predicting a residual for each depth pixel.
We briefly review the core components of PointFlow and the intuition behind our proposed change.

In PointFlow, a set of points called \textit{hypothesis points} are constructed at regular intervals along a depth ray, centered about the depth prediction associated with the given depth ray. The blue and red points in Fig.~\ref{fig:interp} illustrate this.
Features are generated for the hypothesis points.
Then, a network processes these features and outputs a probability score for every point indicating confidence the given point is at the correct depth.
Finally, the expected offset is calculated using these probabilities and added to the original depth prediction.
Our key innovation is the use of our multi-scale scene encoding to generate the hypothesis point features.

In the original PointFlow, hypothesis points are constructed for a \textit{single} depth map, augmented with features using Eq.~\ref{eq:var}, and aggregated into a point cloud.
Note this point cloud is strictly different from our point cloud as (1) it is produced using a \textit{single} depth map, and (2) it includes hypothesis points.
Features are generated for each point using edge convolutions \cite{dgcnn} on the k-Nearest-Neighbor (kNN) graph.
Crucially, these edge convolutions never operate on a unified 3D scene representation in the original PointFlow.
This prevents the offset predictions from learning global information, which we believe is a critical step for depth residual prediction.
Furthermore, because of the required kNN search, this formulation cannot scale to process a joint point cloud from an arbitrary number of depth maps, therefore preventing it from scaling to learn global information.

Inspired by convolutional occupancy networks \cite{peng2020conet} and IFNets \cite{chibane20ifnet}, we instead generate hypothesis features by interpolating each scale of our multi-scale scene encoding (see Fig.~\ref{fig:interp}).
With this key change, we use powerful scene-level priors in our offset prediction conditioned on all $N$ depth predictions for a given scene.
Furthermore, by using the same encoding to update all $N$ depth predictions, we encourage global consistency of predictions.
We now describe in detail our variation of the PointFlow method (see Figs.~\ref{fig:main} and~\ref{fig:interp}), using notation similar to the original paper.

\textbf{Hypothesis Point Construction:} 
For a given back-projected depth pixel $\vect{{p}}$ from depth map $\vect{D}_n$, we generate $2h+1$ point hypotheses $\{\vect{\tilde{p}}_k\}$:
\begin{equation}
    \vect{\tilde{p}}_k = \vect{p} + ks\vect{t}, \quad k = -h, \dots, h
\end{equation}
where $\vect{t}$ is the normalized reference camera direction of $\vect{D}_n$, and $s$ is the displacement step size.

\textbf{Feature Generation:} We generate a multi-scale feature for each hypothesis point $\vect{\tilde{p}}_k$ using trilinear interpolation to point $\vect{\tilde{p}}_k$ of our sparse features volumes $\vect{V}_1, \vect{V}_2, \vect{V}_3$, using $0$s where features are not defined:
\begin{equation}
    \vect{f}_i(\vect{\tilde{p}}_k) = \textrm{sparse\_interp}(\vect{V}_i, \vect{\tilde{p}}_k), \quad i = 1, 2, 3
\end{equation}
Next, we generate a variance feature $\boldsymbol{\sigma}^2(\vect{\tilde{p}}_k)$ for hypothesis point $\vect{\tilde{p}}_k$ using Eq.~\ref{eq:var}.
The final feature for a hypothesis point is the channel-wise concatenation of these features:
\begin{equation} \label{eq:pf-feats}
    \vect{f}_k(\vect{\tilde{p}}_k) = \textrm{concat}\left[ \vect{f}_1(\vect{\tilde{p}}_k), \vect{f}_2(\vect{\tilde{p}}_k), \vect{f}_3(\vect{\tilde{p}}_k), \boldsymbol{\sigma}^2(\vect{\tilde{p}}_k) \right]
\end{equation}
We stack our $2h+1$ point-hypothesis features to form a 2D feature $\vect{H} \in \mathbb{R}^{(2h+1)\times c}$, where $c$ is the sum of the dimensions of our variance and scene encoding features.

\textbf{Offset Prediction:} We apply a 4 layer 1D CNN followed by a softmax function to predict a probability scalar for each point-wise entry in $\vect{H}$. 
The predicted displacement of point $\vect{p}$ is then as follows:
\begin{equation}
    \Delta d_p = \mathbb{E}(ks) = \sum_{k=-h}^{h} ks \times \textrm{Prob}(\vect{\tilde{p}}_k)
\end{equation}
The updated depth for each depth map is the depth of point $\vect{p} + \vect{t}\Delta d_p$ with respect to the camera associated with $\vect{D}_n$. 

\subsection{Bringing It All Together: 3DVNet} \label{sec:3dvnet}

In this section, we describe our full depth-prediction pipeline using our multi-scale volumetric scene modeling and PointFlow-based refinement, which we name 3DVNet
(see Fig.~\ref{fig:3dvnet}).
Our pipeline consists of (1) initial feature extraction and depth prediction, (2) scene modeling and refinement, and (3) upsampling of our refined depth map to the size of the original image.
The scene modeling and refinement is done in a nested for-loop fashion, extracting a scene model in the outer loop and iteratively refining the depth predictions using that scene model in the inner loop.
We fix the input image size of 3DVNet to $320 \times 256$.

\textbf{2D Feature Extraction:} For our 2D feature extraction, we adopt the approach of D\"uz\c{c}eker \etal~\cite{duzceker2020dvmvs}, and use a 32 channel feature pyramid network (FPN) \cite{lin2017fpn} constructed on a MnasNet \cite{tan2019mnas} backbone to extract coarse and fine resolution feature maps of size $80 \times 64$ and $160\times 128$ respectively.
For every image $\vect{I}_n$, we denote these $\vect{F}_n^c$ and $\vect{F}_n^f$.

\textbf{MVSNet Prediction:} For the coarse depth prediction of image $\vect{I}_n$, we use a small MVSNet \cite{yao2018mvs} using the reference and source coarse feature maps $\{\vect{F}^\textit{c}_s : s \in \vect{S}_n\}$ to predict an initial coarse depth $\vect{D}_n^0$.
Our cost volume is constructed using traditional plane sweep stereo with $L = 96$ depth hypotheses sampled uniformly at intervals of 5 cm starting at 50 cm.
Similar to Yu and Gao $\cite{yu2020fmvs}$, our predicted depth map is spatially sparse compared to feature map $\vect{F}^\textit{c}_n$.
We fix our coarse depth map prediction size to $56 \times 56$.

\textbf{Nested For-Loop Refinement:} We denote the updated depths after scene-modeling iteration $l_o$ and PointFlow iteration $l_i$ as $\{\vect{D}_n^{(l_o, l_i)}\}$.
We use initial depth predictions $\{\vect{D}_n^0\}$ and coarse feature maps $\{\vect{F}_n^\textit{c}\}$ to generate multi-scale scene encoding $\vect{V}_1$, $\vect{V}_2$, $\vect{V}_3$.
We then run PointFlow refinement three times with displacement step size $s = 5$ cm, $5$ cm, and $2.5$ cm and $h = 3$ to get updated depths $\{\vect{D}_n^{(1, 3)}\}$.
In early experiments, we found two iterations at $5$ cm to be helpful.
We re-generate our scene encoding using updated depths $\{\vect{D}_n^{(1, 3)}\}$ and coarse feature maps $\{\vect{F}_n^\textit{c}\}$.
We then run PointFlow three more times with step sizes $s = 5$~cm, $5$~cm, and $2.5$ cm and $h = 3$ to get updated depths $\{\vect{D}_n^{(2, 3)}\}$.
We find our depth maps converge at this point.

\textbf{Coarse-to-Fine Upsampling:} In this stage, we upsample each refined depth prediction $\vect{D}_n^{(2, 3)}$ to the size of image $\vect{I}_n$.
We find PointFlow refinement does not remove interpolation artifacts, as this generally requires predicting large offsets across depth boundaries.
We outline a simple, coarse-to-fine method for upsampling while removing artifacts.
See the right section of Fig.~\ref{fig:3dvnet}.
At each step, we upsample the current depth prediction using nearest-neighbor interpolation to the size of the next-largest feature map and concatenate, using the original image $\vect{I}_n$ in the final step.
We then pass the concatenated feature map and depth through a smoothing network.
We use a version of the propagation network proposed by Yu and Gao \cite{yu2020fmvs}.
For every pixel $\vect{p}$ in depth map $\vect{D}$, the smoothed depth $\vect{\tilde{D}}$ is a weighted sum of $\vect{D}$ in the $3\times 3$ neighborhood about $\vect{p}$:
\begin{equation}
    \vect{\tilde{D}}(\vect{p}) = \sum_{\vect{q} \in [-1, 0, 1]^2} g_{\theta}\left(\vect{p},\vect{q}\right)\vect{D}(\vect{p} + \vect{q})
\end{equation}
where $g_{\theta}$ is a 4 layer CNN that takes as input the concatenated feature and depth map and outputs 9 weights for every pixel $\vect{p}$,
and $g_{\theta}\left(\vect{p},\vect{q}\right)$ indexes those weights for the pixel $\vect{p}$.
A softmax function is applied to the weights for normalization.
We apply this coarse-to-fine upsampling to every refined depth map $\{\vect{D}_n^{(2, 3)}\}$ to arrive at a final depth prediction $\{\vect{D}_n\}$ for every input image $\{\vect{I}_n\}$.

\section{Experiments}

\begin{table*}[ht]
\begin{small}
\begin{center}
\begin{tabular}{r c c c c c c c c|c c|c}
\thickhline
 & \multirow{2}{*}{PMVS} & PMVS & \multirow{2}{*}{FMVS} & FMVS & DVMVS & DVMVS & \multirow{2}{*}{GPMVS} & GPMVS & \multirow{2}{*}{Atlas} & Neural- & \multirow{2}{*}{Ours} \\
 & & (FT) & & (FT) & pair & fusion & & (FT) & & Recon & \\
\thickhline 
\sc{\textbf{ScanNet}} & & & & & & & & & & & \\
Abs-rel $\downarrow$ & 0.389 & 0.085 & 0.274 & 0.084 & 0.069 & \underline{0.061} & 0.121 & 0.062 & 0.062 & 0.063 & \textbf{0.040} \\
Abs-diff $\downarrow$ & 0.668 & 0.168 & 0.444 & 0.165 & 0.142 & 0.127 & 0.214 & 0.124 & 0.116 & \underline{0.099} & \textbf{0.079} \\
Abs-inv $\downarrow$ & 0.148 & 0.048 & 0.145 & 0.050 & 0.044 & \underline{0.038} & 0.066 & 0.039 & 0.044 & 0.039 & \textbf{0.026} \\
Sq-rel $\downarrow$ & 0.798 & 0.046 & 0.463 & 0.045 & 0.026 & \underline{0.021} & 0.860 & 0.022 & 0.040 & 0.039 & \textbf{0.015} \\
RMSE $\downarrow$ & 1.051 & 0.267 & 0.776 & 0.267 & 0.220 & 0.200 & 0.339 & \underline{0.199} & 0.238 & 0.206 & \textbf{0.154} \\
$\delta < 1.25$ $\uparrow$ & 0.630 & 0.922 & 0.732 & 0.922 & 0.949 & \underline{0.963} & 0.890 & 0.960 & 0.935 & 0.948 & \textbf{0.975} \\
$\delta < 1.25^2$ $\uparrow$ & 0.768 & 0.981 & 0.857 & 0.979 & 0.989 & \textbf{0.992} & 0.971 & \textbf{0.992} & 0.971 & 0.976 & \textbf{0.992} \\
$\delta < 1.25^3$ $\uparrow$ & 0.859 & 0.994 & 0.915 & 0.993 & \underline{0.997} & \underline{0.997} & 0.990 & \textbf{0.998} & 0.985 & 0.989 & \underline{0.997} \\
\hline
\rowcolor{Gray}
Acc $\downarrow$ & 0.093 & \textbf{0.039} & 0.059 & \underline{0.043} & 0.059 & 0.067 & 0.077 & 0.057 & 0.078 & 0.058 & 0.051 \\
\rowcolor{Gray}
Comp $\downarrow$ & 0.303 & 0.256 & 0.184 & 0.212 & 0.145 & 0.128 & 0.150 & 0.111 & \underline{0.097} & 0.108 & \textbf{0.075} \\
\rowcolor{Gray}
Prec $\uparrow$ & 0.651 & \textbf{0.738} & 0.570 & 0.707 & 0.595 & 0.557 & 0.486 & 0.604 & 0.607 & 0.636 & \underline{0.715} \\
\rowcolor{Gray}
Rec $\uparrow$ & 0.317 & 0.433 & 0.486 & 0.454 & 0.489 & 0.504 & 0.453 & \underline{0.565} & 0.546 & 0.509 & \textbf{0.625} \\
\rowcolor{Gray}
F-score $\uparrow$ & 0.409 & 0.529 & 0.511 & 0.541 & 0.524 & 0.520 & 0.459 & \underline{0.574} & 0.573 & 0.564 & \textbf{0.665} \\
\thickhline
\sc{\textbf{TUM-RGBD}} & & & & & & & & & & & \\
Abs-rel $\downarrow$ & 0.318 & 0.111 & 0.273 & 0.113 & 0.117 & 0.095 & 0.102 & \underline{0.093} & 0.163 & 0.106 & \textbf{0.076} \\
Abs-diff $\downarrow$ & 0.642 & 0.275 & 0.573 & 0.281 & 0.339 & 0.273 & 0.243 & 0.239 & 0.404 & \textbf{0.167} & \underline{0.210} \\
$\delta < 1.25$ $\uparrow$ & 0.662 & 0.858 & 0.694 & 0.851 & 0.838 & 0.886 & 0.874 & 0.891 & 0.816 & \textbf{0.912} & \textbf{0.912} \\
\hline
\rowcolor{Gray}
F-score $\uparrow$ & 0.115 & 0.145 & 0.150 & 0.154 & 0.141 & 0.162 & 0.157 & \underline{0.170} & 0.129 & 0.117 & \textbf{0.181} \\
\thickhline
\sc{\textbf{ICL-NUIM}} & & & & & & & & & & & \\
Abs-rel $\downarrow$ & 0.614 & 0.107 & 0.303 & 0.095 & 0.106 & 0.114 & 0.107 & \underline{0.066} & 0.110 & 0.123 & \textbf{0.050} \\
Abs-diff $\downarrow$ & 1.469 & 0.262 & 0.707 & 0.245 & 0.278 & 0.322 & 0.290 & \underline{0.176} & 0.332 & 0.303 & \textbf{0.120} \\
$\delta < 1.25$ $\uparrow$ & 0.311 & 0.877 & 0.659 & 0.894 & 0.878 & 0.847 & 0.855 & \underline{0.965} & 0.833 & 0.709 & \textbf{0.980} \\
\hline
\rowcolor{Gray}
F-score $\uparrow$ & 0.064 & 0.144 & \underline{0.382} & 0.246 & 0.173 & 0.150 & 0.241 & 0.323 & 0.194 & 0.055 & \textbf{0.440} \\
\thickhline
\end{tabular}
\end{center}
\vspace{-0.6cm}
\caption{Metrics for three datasets (ScanNet, TUM-RGBD, and ICL-NUIM). Bold indicates best performing method, underline the second best. White rows indicate 2D depth metrics while gray rows indicate 3D metrics. Vertical lines separate depth-based methods, volumetric methods, and our method. ``FT" denotes method was finetuned on ScanNet. Our method outperforms all other baseline methods by a wide margin on most metrics.}
\label{tab:main-results}
\end{small}
\end{table*}

\subsection{Implementation and Training Details}

\textbf{Libraries:} Our model is implemented in PyTorch using PyTorch Lightning \cite{falcon2019pl} and PyTorch Geometric \cite{FeyLenssen2019pygeo}.
We use Minkowski Engine \cite{choy2019minkowski} as our sparse tensor library. 
We use Open3D~\cite{Zhou2018o3d} for both visualization and evaluation.

\textbf{Training Parameters:} We train our network on a single NVIDIA RTX 3090 GPU.
Our network is trained end-to-end with a mini-batch size of 2.
Each mini-batch consists of 7 images for depth prediction.
For our loss function, we accumulate the average $L_1$ error between ground truth and predicted depth maps, appropriately downsampling the ground truth depth map to the correct resolution, for all predicted, refined, and upsampled depth map at every stage in our pipeline.
Additionally, we employ random geometric scale augmentation with a factor selected between $0.9$ to $1.1$ and random rotation about the gravitational axis.

We first train with the pre-trained MnasNet backbone frozen using the Adam optimizer \cite{kingma2017adam} with an initial learning rate of $10^{-3}$ which is divided by 10 every 100 epochs ($\sim$1.5k iterations), to convergence ($\sim$1.8k iterations).
We unfreeze the MnasNet backbone and finetune the entire network using Adam and an initial learning rate of $10^{-4}$ that is halved every 50 epochs to convergence ($\sim$1.8k iterations).

\subsection{Datasets, Baselines, Metrics, and Protocols}

\textbf{Datasets:} To train and validate our model, we use the ScanNet \cite{dai2017scannet} official training and validation splits.
For our main comparison experiment, we use the ScanNet official test set, which consists of 100 test scenes in a variety of indoor settings.
To evaluate the generalization ability of our model, we select 10 sequences from TUM-RGBD \cite{sturm2012tum-rgbd}, and 4 sequences from ICL-NUIM \cite{handa2014icl-nuim} for comparison.

\textbf{Baselines:} We compare our method to seven state of the art baselines: Point-MVSNet (PMVS) \cite{chen2019ptmvs}, Fast-MVSNet (FMVS) \cite{yu2020fmvs}, DeepVideoMVS pair/fusion networks (DVMVS pair/fusion) \cite{duzceker2020dvmvs}, GPMVS batched \cite{hou2019gpmvs}, Atlas \cite{murez2020atlas}, and NeuralRecon \cite{sun2021neucon}.
The first five baselines are depth-prediction methods while the last two are volumetric methods.
Of these, we consider GPMVS and Atlas the most relevant depth and volumetric methods respectively, as both use information from all frames simultaneously during inference.
We use the ScanNet training scenes to fintetune methods not trained on ScanNet~\cite{chen2019ptmvs, hou2019gpmvs, yu2020fmvs}.
We report both the finetuned and pretrained results, denoted with and without ``FT".
To account for range differences between the DTU dataset~\cite{jensen2014large} and ScanNet, we use our model's plane sweep parameters with PMVS and FMVS.

\textbf{Metrics:} We use the 2D and 3D metrics presented by Murez \etal~\cite{murez2020atlas} for evaluation.
See supplementary for definitions.
Amongst these metrics, we consider Abs-rel, Abs-diff, and the first inlier ratio metric $\delta < 1.25$ as the most suitable 2D metrics for measuring depth prediction quality, and F-score as the most suitable 3D metric for measuring 3D reconstruction quality.
Following D\"uz\c{c}eker \etal~\cite{duzceker2020dvmvs}, we only consider ground truth depth values greater than 50 cm to account for some methods not being able to predict smaller depth.
We note F-score, Precision, and Recall are calculated per-scene and then averaged across all the scenes.
This results in a different F-score than when calculating from the averaged Precision and Recall reported.

\textbf{Protocols:} For depth-based methods, we fuse predicted depths using the standard multi-view consistency based point cloud fusion.
Based on results on validation sets, we modify the implementation of Galliani \etal \cite{galliani2015gaupuma} to use \textit{depth}-based multi-view consistency check, rather than a \textit{disparity}-based check (see Sec.~3.3 of the supplementary materials).
For volumetric methods, we use marching cubes to extract a mesh from the predicted TSDF.
Following Murez \etal~\cite{murez2020atlas}, we trim the meshes to remove geometry not observed in the ground truth camera frustums.
Additionally, ScanNet ground truth meshes often contain holes in observed regions.
We mask out these holes for all baselines to avoid false penalization.
All meshes are single layer to match ScanNet ground truth as noted by Sun \etal \cite{sun2021neucon}.

We use the DVMVS keyframe selection.
For depth-based methods, we use each keyframe as a reference image for depth prediction.
We use the 2 previous and 2 next keyframes as source images (4 source images total).
For depth-based methods, we resize the output depth map to $640 \times 480$ using nearest-neighbor interpolation.
For volumetric methods, we use the predicted mesh to render $640 \times 480$ depth maps for each keyframe.

\subsection{Evaluation Results and Discussion}

\begin{figure*}[t]
\begin{center}
   \includegraphics[width=0.9\linewidth]{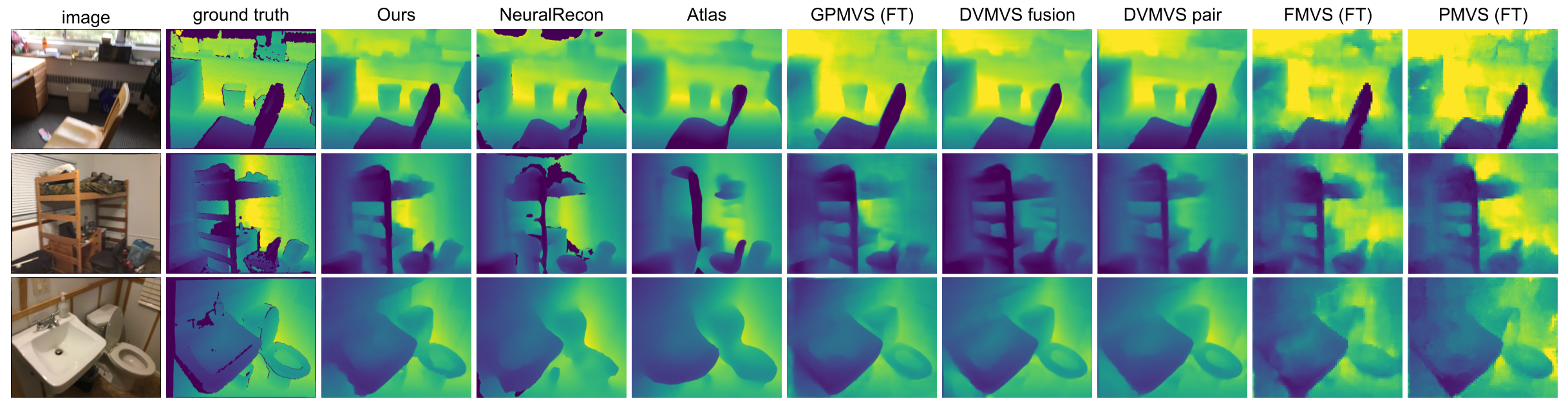}
\end{center}
\vspace{-0.6cm}
\caption{Qualitative depth results on ScanNet. Our method produces sharp details with well defined object boundaries.}
\label{fig:depth-grid}
\end{figure*}

\begin{figure*}[t]
\begin{center}
   \includegraphics[width=0.9\linewidth]{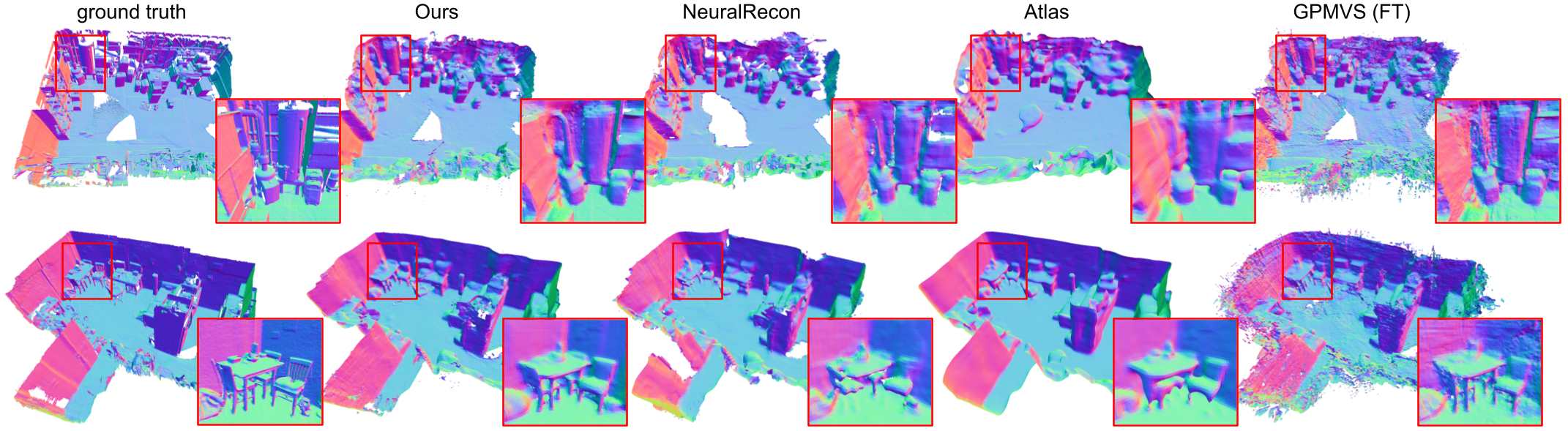}
\end{center}
\vspace{-0.6cm}
\caption{Qualitative reconstruction results on ScanNet for the four best-performing methods. Our technique produces globally coherent reconstructions like purely volumetric methods while containing the local detail of depth-based methods.}
\label{fig:rec-grid}
\end{figure*}

See Tab.~\ref{tab:main-results} for 2D depth and 3D geometry metrics on all datasets.
Our method outperforms all baselines by a wide margin on most metrics.
Notably, our Abs-rel error on ScanNet is 0.021 less than the DVMVS fusion, the second best method, while the Abs-rel of the third, fourth, and fifth best methods are all within 0.002 of DVMVS fusion.
Similarly, Our ScanNet F-score is 0.09 more than GPMVS (FT), the second best method, while the F-score is within 0.001 of GPMVS (FT) for Atlas, the third best method. 
This demonstrates the significant quantitative increase in both depth and reconstruction metrics of our method. Results on additional datasets show the strong generalization ability of our model.

We include qualitative results on ScanNet.
See Figs.~\ref{fig:depth-grid} and~\ref{fig:rec-grid}.
See Sec.~4 of the supplementary materials for additional qualitative results.
Our depth maps are visually pleasing, with clearly defined edges.
They are comparable in quality to those of GPMVS and DVMVS fusion while being quantitatively more accurate.
Our reconstructions are coherent like volumetric methods, without the noise present in other depth-based reconstructions, which we believe is a result of our volumetric scene encoding and refinement.

We do note one benefit of Atlas is its ability to fill large unobserved holes.
Though not reflected in the metrics, this leads to qualitatively more complete reconstructions.
Our system relies on depth maps and thus cannot do this as designed.
However, as a result of averaging across image features, Atlas produces meshes that are overly smooth and lack detail.
In contrast, our reconstructions contain sharper, better defined features than purely volumetric methods.
Finally, we note our system cannot naturally be run in an online fashion, requiring availability of all frames prior to use.

\begin{table}[t]
\begin{small}
\begin{center}
\begin{tabular}{c|c|c c c|g}
\thickhline
$l_o$ & $l_i$ & Abs-rel & Abs-diff & $\delta < 1.25$ & F-score\\
\hline
0 & 0 & 0.070 & 0.137 & 0.949 & 0.559 \\
\hline
1 & 1 & 0.050 & 0.100 & 0.965 & 0.651 \\
1 & 2 & 0.044 & 0.088 & 0.971 & 0.661 \\
1 & 3 & 0.043 & 0.086 & 0.972 & 0.664 \\
\hline
2 & 1 & 0.041 & 0.081 & 0.974 & 0.668 \\
2 & 2 & 0.040 & 0.079 & 0.975 & 0.667 \\
2 & 3 & 0.040 & 0.079 & 0.975 & 0.665 \\
\hline
\end{tabular}
\end{center}
\vspace{-0.6cm}
\caption{Metrics as a function of number of inner PointFlow-refinement iterations (denoted $l_i$) and number of outer-loop scene-modeling passes (denoted $l_o$).}
\label{tab:iters}
\end{small}
\end{table}

\begin{table}[t]
\begin{small}
\begin{center}
\begin{tabular}{c c c c|g}
\thickhline
Model & Abs-rel & Abs-diff & $\delta < 1.25$ & F-score\\
\hline
no 3d & 0.067 & 0.134 & 0.952 & 0.551 \\
single scale & 0.041 & 0.080 & 0.973 & 0.662 \\
avg feats & 0.043 & 0.082 & 0.975 & 0.656 \\
full & 0.040 & 0.079 & 0.975 & 0.665 \\
\hline
\end{tabular}
\end{center}
\vspace{-0.6cm}
\caption{Metrics for our ablation study. See Sec.~\ref{sec:additional-studies} for descriptions of each condition.}
\label{tab:ablation}
\end{small}
\end{table}

\subsection{Ablation and Additional Studies} \label{sec:additional-studies}
\textbf{Does Iterative Refinement Help?} We study the effect of each inner and outer loop iteration of our depth refinement. See Tab.~\ref{tab:iters}.
We exceed state-of-the-art metrics after 2 iterations.
3 additional iterations add continued improvement, confirming the effectiveness of iterative refinement.
By 5 iterations, our metrics have converged, with depth stabilizing and F-score decreasing slightly.
Interestingly, the final iteration appears slightly detrimental.

\textbf{Does Multi-Scale Scene Modeling Help?} To test this, we (1) completely remove our multi-scale scene encoding from the PointFlow refinement, and (2) only use the coarsest scale $\vect{V}_3$, respectively denoted ``no 3D" and ``single scale" in Tab.~\ref{tab:ablation}.
Without any scene-level information, our refinement breaks down, indicating the scene modeling is essential.
The single scale model does slightly worse, confirming the effectiveness of our multi-scale encoding.

\textbf{Do Multi-View Matching Features Help?}
We use a per-channel average instead of variance aggregation for each point in our feature-rich point cloud, denoted ``avg feats" in Tab.~\ref{tab:ablation}.
Most metrics, notably the F-score, suffer.
This supports our hypothesis that multi-view matching is more beneficial for reconstruction than averaging. 

For additional studies, see the supplementary material.

\section{Conclusion}

We present 3DVNet, which uses the advantages of both depth-based and volumetric MVS.
We perform experiments with 3DVNet to show depth-based iterative refinement and multi-view matching combined with volumetric scene modeling greatly improves both depth-prediction \textit{and} reconstruction metrics.
We believe our 3D scene-modeling network bridges an important gap between depth prediction, image feature aggregation, and volumetric scene modeling and has applications far beyond depth-residual prediction.
In future work, we will explore its use for segmentation, normal estimation, and direct TSDF prediction.

\vspace{0.2cm}
\noindent\textbf{Acknowledgements: }Support for this work was provided by ONR grants N00014-19-1-2553 and N00174-19-1-0024, as well as NSF grant 1911230.

{\small
\bibliographystyle{ieee_fullname}
\bibliography{egbib}
}

\end{document}